\documentclass{article}
\usepackage{times}
\usepackage{helvet}
\usepackage{courier}
\usepackage{alltt}
\usepackage{xspace}

\usepackage{amsmath}
\usepackage{amssymb}
\usepackage{graphicx}
\usepackage{algorithm}
\usepackage{algpseudocode}
\usepackage{url}
\usepackage{color}

\mathchardef\mhyphen="2D

\newcommand{\LTL}{\ensuremath{\mathit{LTL}}\xspace}

\newcommand{\Unt}{{\cal U}}

\newcommand{\F}{{\cal F}}
\newcommand{\G}{{\cal G}}

\title{Expressibility of Norms in Temporal Logic}

\author{Natasha Alechina\\ University of Nottingham, UK \\ \texttt{nza@cs.nott.ac.uk} \and 
Mehdi Dastani \\ University of Utrecht, NL\\ \texttt{m.m.dastani@uu.nl} \and 
Brian Logan\\ University of Nottingham, UK \\ \texttt{bsl@cs.nott.ac.uk}}

\date{}

\begin{document}

\maketitle

\begin{abstract}
In this short note we address the issue of expressing norms (such as 
obligations and prohibitions) in temporal logic. In particular, we address
the argument from \cite{Guido} that norms cannot be expressed in Linear Time
Temporal Logic (LTL).
\end{abstract}


Recently, doubts were raised in \cite{Guido} regarding suitability of \LTL and other temporal
logics for expressing `real life' norms. We claim that the argument in
\cite{Guido} does not show that LTL is unsuitable for expressing norms.
Instead it shows that a translation of deontic notions such as
obligations and permissions into temporal logic which interprets `obligatory'
as `always true' and `permitted' as `eventually true' does not work, as could
be expected. However, \cite{Guido} is now often cited as an argument against
using temporal logic for specifying norms in general.
We would like to revisit the example which is
considered paradoxical when specified in \LTL in \cite{Guido}, and show that it is possible
to exactly specify the set of conditions on runs which satisfy
the norms from the example using standard \LTL.

We assume that the reader is familiar with temporal logic, in particular
with the syntax and semantics of \LTL. We use $\Unt$ for the Until operator,
$\G$ for always in the future, $\F$ for some time in the future.

The example is as follows (we compress it slightly without changing the meaning, and use the
same variable names for propositions):

\begin{enumerate}
\item collection of personal information ($A$) is forbidden unless authorised by the court
($C$)
\item The destruction of personal information collected illegally before accessing it ($B$)
excuses the illegal collection
\item collection of medical information ($D$) is forbidden unless collection of personal
information is permitted
\end{enumerate}

As pointed out in \cite{Guido}, this classifies possible situations as
compliant and non-compliant as follows:
\begin{itemize}
\item situations satisfying $C$ are compliant
\item situations not satisfying $C$, where $A$ happens but $B$ happens as well, are weakly
compliant
\item situations where $C$ is false, where $A$ happens and $B$ does not, are violations
\item situations not satisfying $C$ where $D$ happens are violations
\item situations not satisfying $C$ but also not satisfying $A$ and $D$ are
compliant
\end{itemize}
The classification above is not very precise, since $A$, $B$, $C$, and $D$
are treated as state properties which are true or false at the same time.
Later in \cite{Guido} a temporal relation between $A$ and $B$
is introduced: if $C$ is false and $A$ happens, then $B$ should happen some
time after that to compensate for the violation of $A$\footnote{It would have perhaps been better not to treat $B$ as a state property, but as a property
of a run, `data not accessed until destroyed', which is expressible as
$\neg Read\, \Unt Destroyed$, but we will stick with the formalisation in
\cite{Guido} to make comparison easier. Another issue is that instead of
requiring $B$ to happen `eventually', in real life there would be some time
limit on when it should happen (such as in the next state).}.

Hence it is very easy to classify runs into compliant or violating in
\LTL:
\begin{itemize}
\item Fully compliant runs: $$\G(C \vee (\neg C \wedge \neg A \wedge \neg D))$$
(everywhere, either there is a court authorisation, or there is no collection
of personal or medical information)
\end{itemize}
\begin{itemize}
\item Weakly compliant runs: $$\F(\neg C \wedge A) \wedge
\G(\neg C \wedge A \rightarrow \F B) \wedge \G(\neg C \rightarrow \neg D)$$
(there is at least one violation of prohibition on collection of personal
information, but each such violation is compensated by $B$ in the future;
there are no violations of prohibition on collecting medical information)
\end{itemize}
Finally, violations are specified as follows:
\begin{itemize}
\item Violating runs:
$$\F(\neg C \wedge (D \vee (A \wedge \neg \F B)))$$
\end{itemize}
Note that the three formulas above define a partition of all possible runs.
Clearly, there is nothing paradoxical in this specification of the set of norms.

For the sake of completeness we reproduce here the formalisation of the same
set of norms in \cite{Guido} and analyse where the paradoxical results come
from.
The set of norms is formalised in \cite{Guido} as follows:
\begin{description}
\item[N1] $\neg C \rightarrow (\neg A \otimes B)$
\item[N2] $C \rightarrow \F A$
\item[N3] $\G \neg A \rightarrow \G \neg D$
\item[N4] $\F A \rightarrow \F D$
\end{description}
{\bf N1} is intended to say that $B$ compensates for a violation $\neg C \wedge
A$. It uses a connective $\otimes$ which was introduced for expressing contrary
to duty obligations. The truth definition for $\otimes$ as given in
\cite{Guido} is
$$TS, \sigma \models \phi \otimes \psi\ \ \mbox{iff}\ \ \forall i\geq 0, TS,
\sigma_i \models \phi \ \ \mbox{or}\ \ \exists j,k: 0 \leq j \leq k,
TS,\sigma_j \models \neg \phi \ \ \mbox{and}\ \ TS, \sigma_k \models \psi$$
where $TS$ is a transition system, and $\sigma$ a run in $TS$. This makes
$\otimes$ equivalent to
$$\G \phi \vee \F(\neg \phi \wedge \F \psi)$$
This condition is similar to our characterisation of weakly compliant runs,
although it is stated as a property which should be true for all runs.
The condition {\bf N2} is one of the really problematic ones. It aims to
say that if $C$ holds, then $A$ is permitted; `permitted' is identified
with `will eventually happen'. It is quite clear that permission of $A$
cannot be expressed as `$A$ will eventually happen'; the two have completely
different meanings. This does not mean that
\LTL cannot be used for specifying norms, it just means that this particular
way of specifying norms in \LTL is inappropriate. {\bf N4} is problematic
in the same way: instead of saying that an occurence of $D$ is not a violation
under the same conditions as when an occurrence of $A$ is not a violation, it
says that if $A$ is going to happen then $D$ is going to happen --
which is again a completely different meaning. {\bf N3} attempts to say that if
$A$ is prohibited then $D$ is prohibited. However, instead it implies that if
$A$ happens (the antecedent $\G \neg A$ is false) then it does not matter
whether $D$ happens (the implication is still true). Given this formalisation,
which is inappropriate in multiple ways, \cite{Guido} produces an example run
where {\bf N1}--{\bf N4} are true and the prohibition on collecting medical
information is violated.
The first two states on this run are $t_1$, $t_2$:
$$t_1 \models  \neg C \wedge A \wedge D$$
$$t_2 \models B$$
which makes it a weakly compliant run as far as violating prohibition of $A$
is concerned, but a non-compliant run as far as violating the prohibition of
$D$ is concerned. With our \LTL specification of the set of norms it is classifid as a violating run since $\F (\neg C \wedge D)$ holds on it. It does
satisfy {\bf N1}--{\bf N4}, but clearly
the problem is with {\bf N1}--{\bf N4} rather than with the intrinsic
difficulty of expressing norm violating patterns in temporal logic.

\bibliographystyle{plain}
\bibliography{references}
\end{document}